\documentclass[final]{cvpr}

\usepackage{times}
\usepackage{epsfig}
\usepackage{graphicx}
\usepackage{amsmath}
\usepackage{amssymb}

\usepackage{comment}
\usepackage{color}

\usepackage{xcolor}
\usepackage{booktabs}
\usepackage{subcaption}
\usepackage{multirow}
\usepackage{algorithm}
\usepackage{algorithmic}
\usepackage{eqparbox}
\usepackage{array}
\usepackage{dsfont}
\usepackage{transparent}
\usepackage{rotating}
\usepackage{pdflscape}

\newcommand{\myparagraph}[1]{\vspace{1pt}\noindent{\bf #1}}

\usepackage[pagebackref=true,breaklinks=true,colorlinks,bookmarks=false]{hyperref}

\def\cvprPaperID{5223}
\def\confYear{CVPR 2021}

\begin{document}

\title{Learning Decision Trees Recurrently Through Communication}

\author{Stephan~Alaniz\textsuperscript{1,2} \hspace{6mm}
Diego~Marcos\textsuperscript{3} \hspace{6mm}
Bernt~Schiele\textsuperscript{2} \hspace{6mm}
Zeynep~Akata\textsuperscript{1,2,4}
\and\textsuperscript{1}{\normalsize University of Tübingen}\\{\color{white}\textsuperscript{1}}{\normalsize Tübingen AI Center}
\and\hspace{-4mm}\textsuperscript{2}{\normalsize MPI for Informatics}\\\hspace{-4mm}{\color{white}\textsuperscript{2}}{\normalsize Saarland Informatics Campus}
\and\hspace{-4mm}\textsuperscript{3}{\normalsize Wageningen University}\\\hspace{-4mm}{\color{white}\textsuperscript{3}}{\normalsize Wageningen, Netherlands}
\and\hspace{-4mm}\textsuperscript{4}{\normalsize MPI for Intelligent Systems}\\\hspace{-4mm}{\color{white}\textsuperscript{4}}{\normalsize Max Planck Campus}
}

\maketitle

\begin{abstract}
\vspace{-2mm}
Integrated interpretability without sacrificing the prediction accuracy of decision making algorithms has the potential of greatly improving their value to the user. 
Instead of assigning a label to an image directly, we propose to learn iterative binary sub-decisions, inducing sparsity and transparency in the decision making process. The key aspect of our model is its ability to build a decision tree whose structure is encoded into the memory representation of a Recurrent Neural Network jointly learned by two models communicating through message passing.
In addition, our model assigns a semantic meaning to each decision in the form of binary attributes, 
providing concise, semantic and relevant rationalizations to the user.
On three benchmark image classification datasets, including the large-scale ImageNet, our model generates human interpretable binary decision sequences explaining the predictions of the network while maintaining state-of-the-art accuracy.
\end{abstract}

\vspace{-4mm}
\section{Introduction}

The decision mechanism of deep Convolutional Neural Networks (CNNs) is often hidden from the user, hindering their employment in critical applications such as health-care, where a thorough understanding of this mechanism may be required.
The aim for analyzing the decision mechanism, i.e. \emph{introspection}, is to reveal the internal process of the decision maker to a machine learning practitioner or user~\cite{park2018multimodal}. 
However, models offering explanations through introspection may result in a performance loss~\cite{gunning2019darpa,maree2004generic}.

Incorporating recent advances in multi-agent communication~\cite{FoersterAFW16}, we formulate the decision process as an iterative decision tree and embed its structure into the memory representation of a Recurrent Neural Network (RNN).
Our model uses message-passing~\cite{HavrylovT17} with discrete symbols from a vocabulary. A tunable parameter controls whether to learn this vocabulary from scratch or to map it to human-understandable attributes assigning a meaning to every decision to improve its interpretability.
Further, encoding the decision tree into the memory of an RNN retains the flexibility and performance of CNNs while being scalable. Instead of requiring an exponential number of tree nodes with increasing depth, our model learns orders of magnitude fewer nodes with a constant number of model parameters for an arbitrary tree depth. After training, our neural model can be converted exactly into a standard decision tree, being computationally efficient at test time.

\begin{figure}
    \centering
    \includegraphics[width=0.95\linewidth]{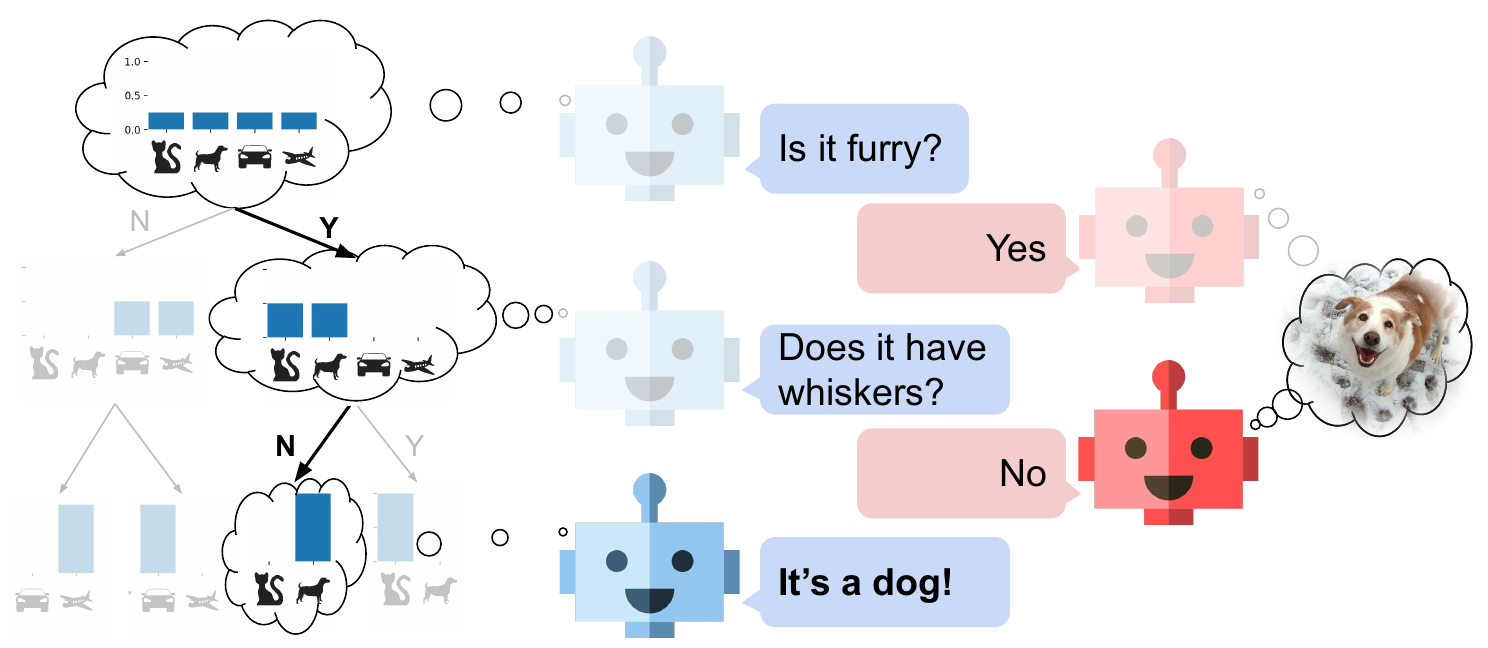}
    \vspace{-2mm}
    \caption{Our \textit{Recurrent Decision Tree (RDT)} (left) asks questions, \textit{Attribute-based Learner (AbL)} (right) answers with a yes/no s.t. the accuracy improves after each step.}
    \vspace{-4mm}
    \label{fig:modelTeaser}
\end{figure}

Our framework (see Figure~\ref{fig:modelTeaser}) exposes a decision path in the form of an explainable decision chain by breaking down the decision process into multiple binary decisions.
Our recurrent decision-tree (RDT) (blue) does not see the image, and has to infer the image class, e.g. \textit{dog}, by recurrently asking binary questions, e.g. \textit{does it have whiskers?} Our attribute-based learner (AbL) (red) answers these questions with yes/no by looking at the image, allowing the RDT to update the class probabilities and the memory representation of the previous questions and answers. This is repeated until the RDT reaches a final decision and the decision tree becomes easily understandable as it associates the binary answers with semantic attributes, e.g. \textit{has whiskers}.

Our contributions are: 1) We propose a recurrent decision tree model (RDTC) with hard node splits and overcome current limitations of decision trees in terms of depth scalability and flexibility; 2) We predict attributes in an end-to-end manner allowing human-interpretable explanations; 3) We showcase on three datasets that our model generates explainable decision trees more efficiently than related methods while retaining the performance of non-explainable CNNs.
Our code is publicly available at: \url{https://github.com/ExplainableML/rdtc}.

\section{Related Work}

\myparagraph{Decision Trees with Neural Networks.}
Decision trees are used across many machine learning tasks and applications, including medical diagnosis~\cite{azar2013decision,kononenko2001machine}, remote sensing~\cite{friedl1997decision,hansen2000global} and judicial decision making~\cite{kleinberg2018human}. They make no assumptions on the data, and are inherently interpretable~\cite{huysmans2011empirical}.

To improve their performance, combining decision trees with neural networks has been explored by building hierarchical classifiers~\cite{Murthy16, Brust20, ZhangCWZ17a, ZhangCWZ17b}, by transferring models~\cite{Humbird19, Siu19, Frosst17,hinton2015distilling}, and through regularization~\cite{wu2018beyond}. 
Recently, \cite{KontschiederFCB16,Tanno18,wan2020nbdt} have proposed learning decision trees directly with neural networks.
NBDT~\cite{wan2020nbdt} constructs trees in the weight space of a neural network and Adaptive Neural Trees~\cite{Tanno18} directly model the neural network as a decision tree, where each node and edge correspond to one or more network modules. The prior work closest to ours is the dNDF~\cite{KontschiederFCB16}, which first uses a CNN to determine the routing probabilities on each node and then combines nodes to an ensemble of decision trees that jointly make the prediction. Our method differs in that 1) we focus on explainability by explicitly only considering a hard binary decision and 2) the depth and branching structure of our decision trees is learned by an RNN instead of being fixed a priori.

\myparagraph{Multi-Agent Communication.}
Learning to communicate in a multi-agent setting has gained interest with the emergence of deep reinforcement learning \cite{FoersterAFW16,HavrylovT17,LazaridouHTC18,CaoLLLTC18,JiangL18,DasGRBPRP19,Corona19, LazaridouPT20, LazaridouM20}.
Most works focus on establishing a novel communication protocol from scratch. \cite{FoersterAFW16} and \cite{CaoLLLTC18} train multiple agents to maximize a shared utility by establishing their own language. However, large scale multi-agent settings can suffer from too much communication, as valuable information comes with extensive computations~\cite{JiangL18}. Targeted communication focuses on key information and allows iterative exchange of information before performing a task that can improve both performance and interpretability~\cite{DasGRBPRP19}.

Image reference games are used to study the emergence of language~\cite{LazaridouHTC18} and effectiveness in communication also when concepts are being misunderstood~\cite{Corona19}.
\cite{HavrylovT17} propose an agent that composes a message of categorical symbols to another agent that uses the information in these messages to solve a referential game.
Our model in contrast allows both to learn a communication protocol from scratch or use human-understandable concepts as a vocabulary.

\myparagraph{Attributes.} Attributes are human understandable visual properties of objects that are shared between classes. Attributes have been used for image description \cite{FZ08, CGG12}, caption generation \cite{OGB11}, face recognition \cite{CGG13}, image retrieval \cite{KBNFTZ08,SFD11}, action recognition \cite{QW12,YJKLGL11}, novelty detection \cite{WB13} and object classification \cite{Lampert14,SYHX14,MSN11,SQL12, ChenCHRZ19}. In this work we propose to use attributes as explanations, i.e. they label the branches in the learned decision tree,
allowing users to easily inspect the reasoning encoded by the tree.

\myparagraph{Explainability through sparsity.}
Optimizing representations to be sparse~\cite{zhang2016l1} when seeking interpretability~\cite{wright2010sparse} draws some resemblance with the working memory of humans~\cite{ma2014changing}, which is limited to a handful of items at the same time. \cite{doshi2017towards} hypothesizes that the nature of these items (they need to be understandable \emph{per se}), their number and the structure in which they are presented all impact the interpretability of a representation. 
Furthermore, interpretability can be achieved by regularizing neural networks such that their representations, not only to become sparse~\cite{marcos2020contextual}, but also adopt the structure of a decision tree~\cite{wu2018beyond}. 

Although both sparsity and tree depth have been used as proxies for interpretability in decision trees, human studies suggest that the best proxy is problem-dependent~\cite{lage2018human}. 
Beyond explainable ML, a sparse representation is considered to be essential for moving towards hybrid deep learning-symbolic models~\cite{cranmer2020discovering,marcus2020next} and for obtaining representations that are closer to conscious reasoning~\cite{bengio2017consciousness}. Indeed, a recent model of how human brains work postulate a conceptualization step, linked to dimensionality reduction, followed by an attention mechanism that sparsely selects concepts~\cite{cortese2019neural}.

\begin{figure*}[t]
    \centering
    \includegraphics[width=0.9\linewidth]{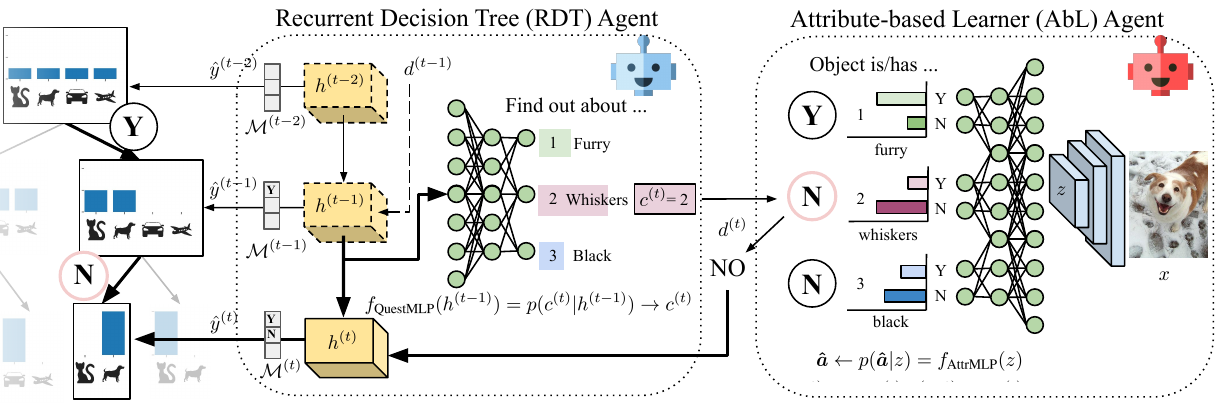}
    \caption{A single communication step between the RDT (left) and AbL (right) in our \texttt{RDTC} framework. RDT uses the hidden state \(h^{(t-1)}\) of its LSTM (yellow) to requests a single attribute \(a_{c^{(t)}}\) by selecting it through its \(f_{\text{QuestMLP}}\). AbL uses its \(f_{\text{AttrMLP}}\) to predict a binary response \(d^{(t)} = \boldsymbol{\hat{a}}_{c^{(t)}}\) indicating the presence/absence of the attribute. Finally, RDT updates its state \(h^{(t)}\) and explicit memory \(\mathcal{M}^{(t)}\) with the binary response to improve its classification prediction \(\hat{y}^{(t)}\).}
    \label{fig:schemaAttribute}
\end{figure*}

\section{\texttt{RDTC} Framework}
\label{sec:model}

Our Recurrent Decision Tree via Communication framework is a sequential interaction between the Recurrent Decision Tree (RDT) and Attribute-based Learner (AbL) models trained to classify images by communicating (see Figure~\ref{fig:schemaAttribute}).
RDT learns a decision path allowing introspection and AbL provides attribute-based rationales to make the communication human-understandable.

\subsection{Communication between RDT and AbL}
For any single image \(x\), our RDT model iteratively aggregates information into an explicit memory \(\mathcal{M}\) that is sufficient to predict the correct class label \(y \in \mathcal{Y}\). Initially, it starts with no prior information  \(\mathcal{M}^{(0)}\). To gather more information, the RDT agent sends a query message \(c^{(t)}\) to the AbL agent.
The AbL answers the query \(c^{(t)}\)
with a binary response \(d^{(t)}  \in \{0, 1\}\) that RDT uses to update its explicit memory \(\mathcal{M}^{(t)} = \mathcal{M}^{(t-1)} \oplus (c^{(t)}, d^{(t)})\) to improve its class prediction.
This constitutes one iteration \(t\) of the agent-to-agent communication. The interaction repeats until a maximum number of steps is reached or until convergence. 

\myparagraph{Communication Protocol.} The vocabulary size \(|A|\) is set to the total number of attributes for every dataset. 
RDT and AbL learn to communicate with the set of tokens provided by the vocabulary in an end-to-end manner. Note that, the AbL agent attaches a human-understandable meaning to these tokens when annotated attribute data is available. 

At each communication step \(t\), 
RDT chooses one attribute \(a_{c^{(t)}}\) from the vocabulary, identified by its index \(c^{(t)}\), and requests its presence or absence in the image. AbL then provides its binary prediction of this attribute, i.e. \(d^{(t)}\).
We deliberately limit the messages of AbL to be binary as clear yes/no answers are easier to interpret.

\myparagraph{Discrete Messages.}
RDT asks for the attribute via the index \(c^{(t)}\) and the AbL responds with a binary \(d^{(t)}\).
The Gumbel-softmax estimator~\cite{JangGP16,MaddisonMT16} allows to sample from a discrete categorical distribution via the reparameterization trick~\cite{kingma2014,rezendeMW14} to obtain the gradients of this sampling process. We sample \(g_i\) from a \(\textrm{Gumbel}\) distribution and then compute a continuous relaxation of the categorical distribution:
\begin{equation}
    \text{GumbelSoftmax}(\log \boldsymbol{\pi})_i = \frac{\exp((\log \pi_i +  g_i)/\tau)}{\sum_{j=1}^K\exp((\log \pi_j +  g_j)/\tau)}
\end{equation}
where \(\log \pi\) are the unnormalized log-probabilities of the categorical distribution, \(\tau\) is the temperature that parameterizes the discrete approximation. When \(\tau \approx 0\), the output is a one-hot vector and otherwise, it is a continuous signal.

Stochasticity is important for exploring all possible indices \(c^{(t)}\) of vocabulary \(A\) to find the most relevant attribute at each step \(t\). Therefore, we use Gumbel-softmax with $K=|A|$ to sample the attribute index \(c^{(t)}\) for RDT.
As each \(d^{(t)}\) corresponds to the presence or absence of an attribute in \(x\), a deterministic prediction is beneficial. 
By introducing a temperature \(\tau\) to a regular softmax~\cite{hinton2015distilling} in AbL, we approximate the \(\arg \max\) function deterministically as \(\tau\) approaches 0:
\begin{equation}
    \text{TempSoftmax}(\log \boldsymbol{\pi})_i = \frac{\exp(\log \pi_i/\tau)}{\sum_{j=1}^K\exp(\log \pi_i/\tau)}
\end{equation}
Since we use binary attributes, in this case $K=2$. Popular training strategies include (a) annealing \(\tau\) over time and (b) augmenting the soft approximation with an \(\arg \max\) that discretizes the activation in the forward pass and results in the identity function in the backward pass. Using (b) guarantees the communication to be always discrete.

\subsection{Recurrent Decision Tree (RDT) Model}

RDT consists of three parts: an explicit memory \(\mathcal{M}\), an LSTM~\cite{Hochreiter97}, and a question-decoder module, \emph{Question MLP} (see Figure~\ref{fig:schemaAttribute} (left)). \(\mathcal{M}^{(t)}\) contains all the binary attributes, i.e. the responses of AbL \(d_{1:t}\) up to step \(t\). The LSTM keeps track of the attribute order with its hidden state \(h^{(t)}\) to encode the current point in the decision tree and decide on the next question.
RDT decodes its last hidden state \(h^{(t-1)}\) into a categorical distribution via \(f_{\text{QuestMLP}}\):
\begin{equation}
    \log p(c^{(t)}|h^{(t-1)}) = f_{\text{QuestMLP}}(h^{(t-1)})
\end{equation}
where \(p(c^{(t)}|h^{(t-1)})\) indicates the likelihood of requesting a particular attribute. We denote the attribute index \(c^{(t)} \in \{1, \dots, |A|\}\) sampled by:
\begin{align}
    c^{(t)} &= \text{GumbelSoftmax}(f_{\text{QuestMLP}}(h^{(t-1)})).
\end{align}

After each iteration of the communication loop, RDT updates its explicit memory \(\mathcal{M}^{(t)} = \mathcal{M}^{(t-1)} \oplus (c^{(t)}, d^{(t)})\). Concretely, \(\mathcal{M} \in \{0, 1\}^{|A| \times 2}\) is initialized with all zeros and at each time step, we set \(\mathcal{M}_{c^{(t)},d^{(t)}} := 1\).
Encoding the attribute in a one-hot vector helps to indicate missing information with all zeros. \(\mathcal{M}^{(t)}\) keeps track of already observed attributes and their values. 
RDT updates $h^{(t)}$ with: 
\begin{equation}
    h^{(t)} = \text{LSTM}(h^{(t-1)}, \mathcal{M}^{(t)}, c^{(t)}, d^{(t)}).
\end{equation}
 and at each time step \(\mathcal{M}\) is used to predict the class label: 
\begin{equation}
    \hat{y}^{(t)} = f_{\text{ClassMLP}}(\mathcal{M}^{(t)}).
\end{equation}

Since the primary objective of RDT is to maximize the classification performance, we minimize the CE loss between the predicted and the true class probabilities:
\begin{equation}
    \mathcal{L} = \frac{1}{T} \sum_{t=1}^{T} \mathcal{L}_{CE}(y, \hat{y}^{(t)}) = - \frac{1}{T} \sum_{t=1}^{T} \sum_i y_i \log\hat{y}^{(t)}_i.
    \label{eq:clsloss}
\end{equation} 
By averaging the $\mathcal{L}_{CE}$ over all \(T\) time steps, RDT predicts the correct class in a small number of communication steps which also allows it to be evaluated at any intermediate step unlike most other decision tree models that classify only at the leaf nodes (see supplementary for a comparison).

Since our decision tree can be evaluated after every communication step, the depth of the tree is not a fixed hyperparameter, but can be adaptively chosen at test time. This provides a flexible model that can be tuned for higher interpretability (shallow tree) or higher performance (deeper tree) at test time without the need for retraining.

\subsection{Attribute-based Learner (AbL) Agent}
\label{sec:bin}
The AbL feeds its CNN image features \(z\) to \(f_{\text{AttrMLP}}\) to predict a set of learned binary attributes queried by the RDT (Figure~\ref{fig:schemaAttribute}, right) where
softmax with temperature gives us binary attributes \(\boldsymbol{\hat{a}} \in \{0, 1\}^{|A|}\), the discretization of \(p(\boldsymbol{\hat{a}}|z)\):
\begin{align}
    \boldsymbol{\hat{a}} &= \text{TempSoftmax}(f_{\text{AttrMLP}}(z)).
\end{align}
When the RDT requests the attribute with the index \(c^{(t)}\), the AbL simply returns the binarized response about the attribute using \(c^{(t)}\), i.e. $ d^{(t)} = \boldsymbol{\hat{a}}_{c^{(t)}}$.
The attributes are either discovered in an end-to-end manner by optimizing the loss in Equation~\ref{eq:clsloss} (\texttt{RDTC}, i.e. Recurrent Decision Tree via Communication) or they are predicted as human-interpretable concepts using an attribute loss (\texttt{aRDTC}, i.e. attribute-based Recurrent Decision Tree via Communication).

\myparagraph{Attribute Loss.} Minimizing the classification loss at each time step is equivalent to finding a binary data split that reduces the class-distribution entropy the most,
i.e. information gain in classical decision trees. However, a split that best separates the data is not always easy to interpret, especially when the features used for this split result from a non-linear transformation as in a CNN.

We propose to integrate further interpretability by learning \(\boldsymbol{\hat{a}}\) that align with class-level human-annotated attributes \(\alpha\) using a second cross-entropy term weighted by \(\lambda\):
\begin{equation}
    \mathcal{L} = \frac{1}{T} \sum_{t=1}^T \Big[(1-\lambda) \mathcal{L}_{CE}(y, \hat{y}^{(t)}) + \lambda \mathcal{L}_{CE}(\alpha_{y,c^{(t)}}, \boldsymbol{\hat{a}}_{c^{(t)}})\Big].
    \label{eq:attrloss}
\end{equation}
Note that the attribute loss is imposed only on those attributes employed by the model. If an attribute is deemed not to be useful, e.g., if an attribute is weak or hard to predict, our RDT model learns to ignore that attribute.

When $\lambda > 0$, our model (aRDTC) learns to use ground-truth attributes and gives the binary splits a semantic meaning. For instance, the question of RDT for attribute with index \(c^{(t)}\) can be interpreted as ``does it have a black beak?'' with \(a_{c^{(t)}}\): ``has black beak''. 
When $\lambda = 0$, RDTC does not use any human-annotated attributes and automatically discovers them.
Either of these settings may be desirable given the application as we show empirically.

\begin{figure}[t]
\begin{minipage}{0.5\textwidth}
\begin{algorithm}[H]
\begin{algorithmic}[1]
    \REQUIRE{Training images \(X\)\\
             \hspace{2.2em}Stopping \textit{threshold}}
    \ENSURE{Decision tree DT}
    \STATE DT = empty decision tree
    \FOR{\(x\) in \(X\)}
        \STATE DT.reset\_to\_root\_node()
        \STATE \(\boldsymbol{\hat{a}} = \text{AbL}(x)\) \COMMENT{attributes from image}
        \FOR{\(t=1\) to \(n\)}
    	    \STATE \(\hat{y}^{(t)}, c^{(t)} = \text{RDT.step}(d^{(t)})\) \COMMENT{class/attribute of node}
    	    \IF{not DT.node\_exists()}
    	        \STATE DT.add\_node(\(\hat{y}^{(t)}, c^{(t)}\))
    	    \ENDIF
    	    \IF{\(\max_i \hat{y}^{(t)}_i >\) \textit{threshold}}
    	        \STATE break \COMMENT{prune when confident}
    	    \ENDIF
    	    \STATE \(d^{(t)} = \boldsymbol{\hat{a}}_{c^{(t)}}\) \COMMENT{attribute yes/no}
    	    \STATE DT.to\_next\_branch(\(d^{(t)}\)) \COMMENT{1 $\rightarrow$ left; 0 $\rightarrow$ right}
        \ENDFOR
    \ENDFOR
    \STATE \textbf{return} DT
\end{algorithmic}
\caption{RDTC decision tree distillation}
\label{alg:test}
\end{algorithm}
\end{minipage}
\end{figure}

\subsection{Decision Tree Distillation}

The RDT and AbL are trained end-to-end since the communication between these two models is differentiable. At test time
we distill the RDT into an explicit decision tree, i.e., the global structure of nodes, including splitting feature and threshold. The distilled decision tree then models the trained neural network $f_\text{RDT} \equiv f_\text{DT}$.

Algorithm~\ref{alg:test} describes the procedure of extracting a decision tree from RDT. The decision nodes of the decison tree make hard splits based on the presence/absence of an attribute.
GumbelSoftmax adds stochasticity to RDT, which is useful for training, but at test time deterministically choosing the attribute with highest probability is essential for improving the performance and learning a static tree. Hence, it is replaced with TempSoftmax.

We start with an empty decision tree and fill it with nodes as we run the training data through the whole RDTC model (lines 1-2). Whenever a previously unseen node is discovered, we add it to the tree including information about the attribute (\(c^{(t)}\)), where the next node is added, i.e., left of the current node if \(d^{(t-1)}\) is 1 or to the right if \(d^{(t-1)}\) is 0 (line 14), and the current prediction of the class labels \(\hat{y}^{(t)}\) (lines 7-9). We prune the distilled decision tree, i.e. we stop adding nodes to the tree once \(\hat{y}\) is greater than a threshold (=0.95) for a class (lines 10-12). The end result is a decision tree that outputs the same class predictions as our trained neural network RDT given the attribute prediction from our AbL, while being fully explainable by matching learned attributes with human-annotated attribute data.

\section{Experiments}
\myparagraph{Datasets and attributes.} 
We validate our model on the large-scale ImageNet~\cite{ILSVRC15} with 1.2M images from 1K classes. In addition, we use AWA2~\cite{Lampert14, XianLSA19} and CUB~\cite{WahCUB_200_2011}, i.e. two medium-scale benchmark attribute datasets. AWA2 comprises 37K images from 50 animal classes with 85 attributes, while CUB contains 11K images from 200 fine-grained bird species with 312 attributes. Since our model considers splits on hard decisions, we binarize the attributes on all datasets with a threshold at 0.5, i.e., an attribute is present if more than 50\% of the annotations agree. 
When an official classification test set is not provided, for all experiments across the datasets, we randomly assign 20\% of each class as test data and 10\% of the training data as a validation set to tune hyperparameters. 

\myparagraph{Architecture and parameters.} The MLPs consist of two layers with a ReLU non-linearity. We learn the temperature \(\tau\) of the Gumbel-softmax estimator jointly with the network from an initial value for \(\tau\). During training, we always roll out the decision sequence to a maximum number of steps. At test time, we apply our decision tree distillation and stop as soon as the RDT reaches a confidence level specified by a \textit{threshold} parameter (or once the maximum number of decisions is reached). We report the mean per-class accuracy over 5 runs
to avoid bias towards highly populated classes.

\subsection{Comparing with the State of the Art}

We compare our \texttt{aRDTC} and \texttt{RDTC} with classical decision trees (aDT and DT) as baselines, ResNet (ResNet~\cite{He16} and aResNet) and Deep Neural Decision Forests (dNDF)~\cite{KontschiederFCB16} as the state of the art.

\myparagraph{ResNet and aResNet.} ResNet-152 pre-trained on ImageNet and fine-tuned on each of the datasets including its softmax classifier serves as non-explainable deep neural network (ResNet).
Augmented with attribute data, we train aResNet by first predicting the attributes with the same architecture as our AbL model and then a linear layer on top.

\myparagraph{Our \texttt{aRDTC} and \texttt{RDTC}.} Our attribute-based recurrent decision tree (\texttt{aRDTC}) (Section \ref{sec:bin}) uses the attribute loss to associate a human-understandable meaning to the binary decisions.
On the other hand, our recurrent decision tree (\texttt{RDTC}) does not use an attribute loss (\(\lambda = 0\)), and therefore purely optimizes classification performance.

{
\setlength{\tabcolsep}{6pt}
\renewcommand{\arraystretch}{1}
\begin{table}[t]
    \centering
    \begin{tabular}{l c c c}
        Model & \textbf{AWA2} & \textbf{CUB} & \textbf{ImageNet}\\
        \midrule
        ResNet~\cite{He16}  & 98.2 \(\pm\) 0.0 & 79.0 \(\pm\) 0.2 & 73.0 \(\pm\) 0.1\\
        aResNet & 98.3 \(\pm\) 0.0 & 77.3 \(\pm\) 0.5 & N/A \\
        \midrule
        DT & 92.3 \(\pm\) 0.4 & 43.5 \(\pm\) 0.3 & 55.2 \(\pm\) 1.0 \\
        dNDF~\cite{KontschiederFCB16} & 97.6 \(\pm\) 0.2 & 73.8 \(\pm\) 0.3 & 72.6 \(\pm\) 0.1 \\
        \texttt{RDTC} (Ours) & \textbf{98.0} \(\pm\) 0.1 & \textbf{78.1} \(\pm\) 0.2 & \textbf{72.8} \(\pm\) 0.1\\
        \midrule
        aDT & 97.9 \(\pm\) 0.9  & 70.6 \(\pm\) 1.3  & N/A \\
        \texttt{aRDTC} (Ours) & \textbf{98.1} \(\pm\) 0.0 & \textbf{77.9} \(\pm\) 0.6 & N/A
    \end{tabular}%
    \caption{Comparing our \texttt{aRDTC} ($\lambda=0.2$) and \texttt{RDTC} ($\lambda=0$) to the decision tree (aDT and DT), closely related dNDF~\cite{KontschiederFCB16}, and ResNet~\cite{He16} (aResNet, i.e. ResNet with attribute prediction).
    As ImageNet do not have attributes, aResNet, \texttt{aRDTC} and aDT are not applicable (over 5 runs).
    }
    \label{tab:cls_acc}
\end{table}
}

\myparagraph{dNDF.}
The dNDF explicitly models the decision tree by mapping each inner node to an output neuron with sigmoid activation. These nodes define the routing probabilities of the input to the leaves through exhaustive tree traversal where each leaf node stores a class distribution. The final prediction is the averaged class prediction weighted by the routing probabilities of every leaf.
As using multiple randomized trees weakens the interpretability, for a fair comparison, we use a single tree instead of random forests.

\myparagraph{aDT and DT.}
The classical decision tree (DT) is learned on top of the same image features \(z\) by the perceptual module. At each time step, the dataset is split using a single dimension of \(z\)
until a leaf node only contains samples of the same class or a regularization strategy leads to early stopping. 
We incorporate attributes into the DT baseline, i.e. Attribute Decision Tree (aDT).
First, we train a MLP on top of the image features \(z\) to predict class attributes using a binary cross-entropy loss analogously to the attribute loss of our \texttt{aRDTC} model. Second, we fit a decision tree on these predicted attributes for each image to determine the class.
Both DT and aDT are learned using the CART algorithm~\cite{breiman1984classification} and the Gini impurity index as splitting criterion due to its computational advantage over entropy-based methods~\cite{Raileanu2004}.

\myparagraph{Classification results.}
As observed in Table~\ref{tab:cls_acc}, compared to the Decision Tree baselines of their kind, our model variants achieve significantly higher accuracy across all datasets, e.g. \texttt{RDTC} vs DT achieves $98.0$\% vs $92.3$\% and \texttt{aRDTC} vs aDT achieves $98.1$\% vs $97.9$\% on AWA2 because our model scales better and reaches consistent results through gradient-based optimization. Moreover, although \texttt{RDTC} and \texttt{aRDTC} work with constrained single-bit communications to improve explainability, they succeed in maintaining the accuracy of the non-explainable state-of-the-art across all datasets, e.g. $72.8$\% vs $73.0$\% on ImageNet.

Fine-grained decision splits are extremely challenging to explain because objects are visually similar to each other and the distinguishing factor is nuanced. Despite this challenge on CUB, the classification accuracy of \texttt{RDTC} is almost twice as high than classical decision trees that use the same deep features, i.e., $78.1$\% vs $43.5$\% DT.  On the other hand, our \texttt{RDTC} not only outperforms dNDF ($78.1$\% vs $73.8$\%), 
our model exhibits improved interpretability, because we use hard instead of soft binary splits.
As it is hard for non-experts to judge the correctness of the predictions, explanations in this domain are particularly important. Typically, associating a semantic meaning to the decision path improves human interpretability with a significant loss in accuracy, e.g. aResNet vs ResNet ($77.3$\% vs $79.0$\%). On the other hand, on our model this trade-off is less pronounced. When trained with the attribute loss, i.e. \texttt{aRDTC} achieves a higher accuracy compared to aDT ($77.9$\% vs $70.6$\% on CUB) as well as aResNet in addition giving a semantic meaning to the splits.

\begin{figure}[t]
    \centering
    \includegraphics[width=1.\linewidth, trim=0 0 0 0, clip]{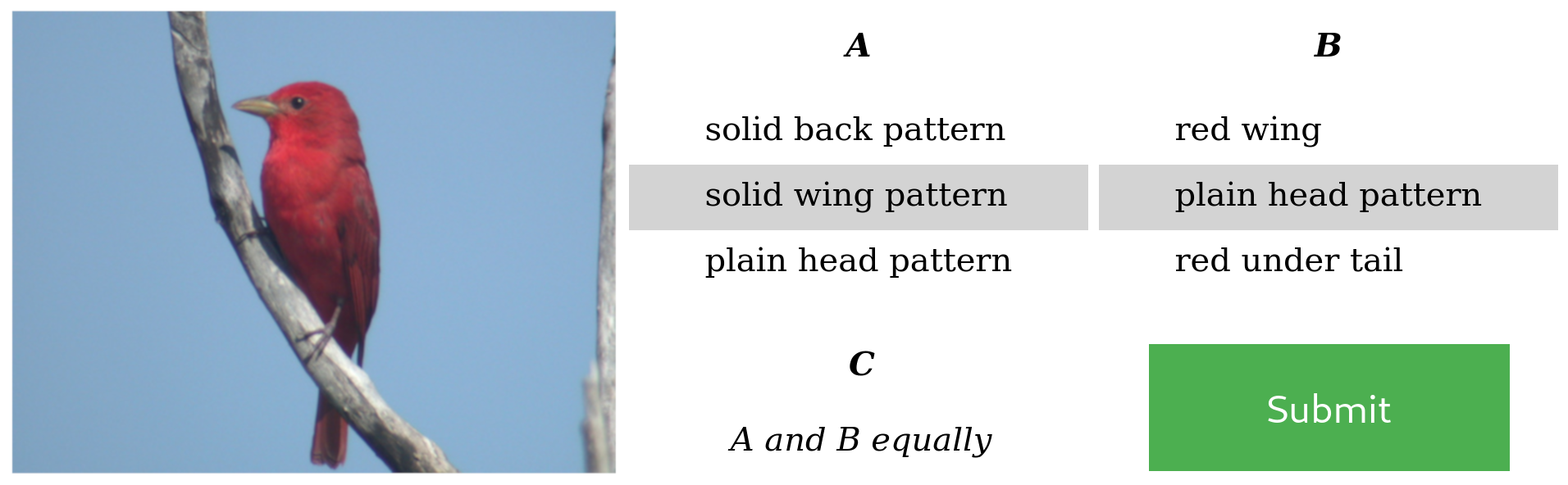}
    \caption{The user picks which set of 3 attributes best fit the image or if they match equally well (attributes come from 2 models out of aRDTC, aDT, aResNet at a time).}
    \label{fig:user_study}
\end{figure}

\begin{figure}[t]
    \centering
    \includegraphics[width=\linewidth, trim=0 0 0 0, clip]{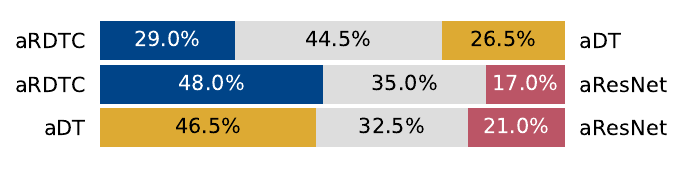}
    \caption{User study results. We show how often the attributes of one model were preferred over any other and when both were found equal (middle).}
    \label{fig:user_study_results}
\end{figure}

\myparagraph{User study.}
The use of named attributes enables humans to understand the decision of the model. However, if all attributes are allowed to be used simultaneously, such as in aResNet, this decision becomes less comprehensible. In contrast, aRDTC provides a sparse solution that considers only a subset of attributes for each prediction.
To quantify the relevance of the selected attributes, we perform a user study with aRDTC, aDT and aResNet on CUB. Since our aRDTC predicts the class label at each step, we select the attributes that change the class probability the most to determine the most critical attributes for the decision. For aDT we use the Mean Decrease Impurity (MDI)~\cite{LouppeWSG13} to find features of maximum importance and for aResNet we select the attributes with the highest weight for the output class. 

The user is prompted with an image as well as two sets of three attributes, i.e., the three most relevant attributes from two models at a time. As some attributes are difficult to recognize, e.g. cone beak, we provide attribute icons with their names and a bird anatomy sketch. The task is to select the set of three attributes that best match the image (see Figure~\ref{fig:user_study}). The user can also report that both sets of attributes fit the image equally well.
We repeat the study on 600 randomly selected images from the CUB test such that each model is compared 200 times against every other model.

We measure how often the attributes of each model are chosen over the other models. Since we only show attributes of two models at a time, we obtain a direct comparison for all pairs of models.
Our results in Figure~\ref{fig:user_study_results} indicate that decision tree models select more relevant attributes than aResNet. The attributes of aRDTC are preferred much more often than aResNet ($48\%$ vs. $17\%$). Similarly, aDT is selected more often than aResNet ($46.5\%$ vs. $21\%$). When comparing the two decision tree models, our aRDTC is slightly favored at $29\%$ over aDT at $26.5\%$ with the majority of users finding them produce equally fitting attributes ($44.5\%$). These results suggest that the tree structure of the decision making also helps in isolating more relevant attributes by putting more weight on individual attributes selected early by the decision tree rather than spreading the contribution among all attributes.

\begin{figure}[t]
\centering
        \includegraphics[width=0.23\textwidth, trim=5 10 5 0,clip]{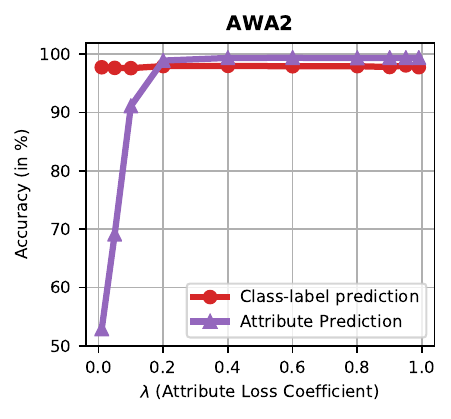}
        \includegraphics[width=0.23\textwidth, trim=5 10 5 0,clip]{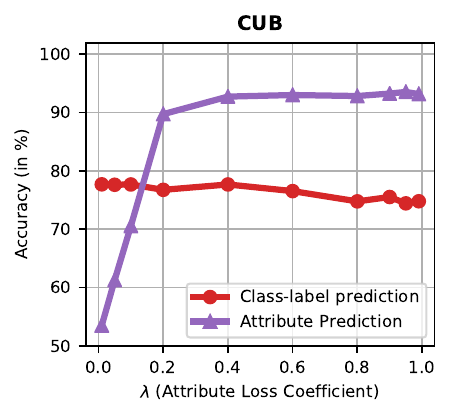}
    \caption{Explainability trade-off with \texttt{aRDTC} on AWA2 and CUB. We vary \(\lambda\) of the attribute loss and report image classification accuracy of RDT (red) and attribute prediction accuracy of AbL (purple). \(\lambda \in [0.01, 0.99]\).}
    \label{fig:tradeoff}
\end{figure}

\subsection{Evaluating The Model Components}

In this section, we evaluate several aspects of our model such as its behavior towards accuracy-explainability tradeoff, ablating its memory mechanism and scalability.

\myparagraph{Accuracy and Explainability Trade-Off}
The trade-off between the classification loss and the attribute loss in our \texttt{aRDTC} model can be measured by varying $\lambda \in [0.01, 0.99]$. 
Our results on AWA2 and CUB in Figure~\ref{fig:tradeoff}
show a slight decrease in the overall classification accuracy (red curve), when $\lambda$ approaches to 1.0 which gives more weight to the attribute prediction as opposed to class label prediction. Indeed, \texttt{RDTC} achieves a higher accuracy than \texttt{aRDTC} that is trained with the attribute loss indicating a tradeoff between explainability and accuracy. Increasing \(\lambda\) leads to a slight decrease in classification accuracy, and generally similar to that of fully optimizing class prediction when \(\lambda=0\).

{
\setlength{\tabcolsep}{4pt}
\renewcommand{\arraystretch}{1}
\begin{table}[t]
    \centering
    \begin{tabular}{l c c c }
        Model & \textbf{AWA2} (\# att)  & \textbf{CUB} (\# att)   & \textbf{ImageNet} (\# att) \\
        \midrule
        \texttt{RDTC-L} & 97.7 (19)  & 73.0 (50)  & 60.8 (159)\\
        \texttt{RDTC-M} & 97.9 (57)  & 77.2 (93)  & 71.6 (82)\\
        \texttt{RDTC}  & \textbf{98.0} (30)  & \textbf{78.1} (42) &  \textbf{72.8} (46)\\
        \midrule
        \texttt{aRDTC-L} & 97.9 (29) & 69.1 (32)  & N/A\\
        \texttt{aRDTC-M} & 98.0 (37)  & 76.4 (52)  & N/A\\
        \texttt{aRDTC}  & \textbf{98.1} (34) & \textbf{77.9} (38)  & N/A
    \end{tabular}%
    \caption{Ablating the memory mechanism of \texttt{aRDTC} ($\lambda=0.2$) and \texttt{RDTC} ($\lambda=0$). Tree state is encoded as either only the LSTM (\texttt{L}), only the explicit memory (\texttt{M}) or both (+ median number of distinct attributes the model learns).
    }
    \label{tab:cls_acc_ablation}
\end{table}
}

\begin{figure*}[t]
    \begin{subfigure}{0.9\textwidth}
        \begin{center}
            \includegraphics[width=0.32\textwidth]{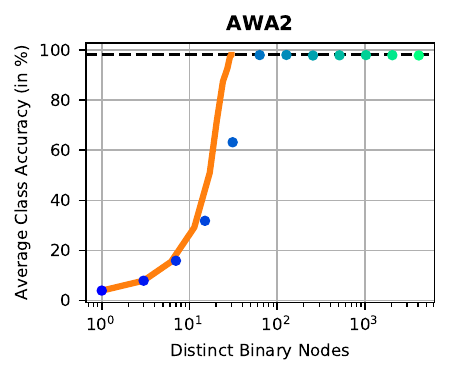}
            \includegraphics[width=0.32\textwidth]{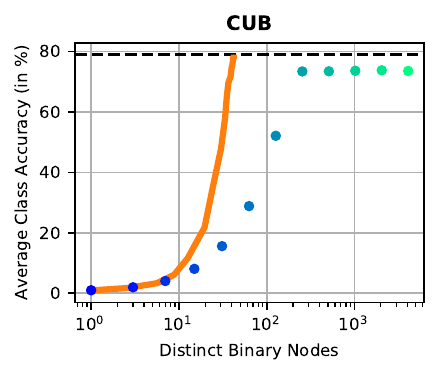}
            \includegraphics[width=0.32\textwidth]{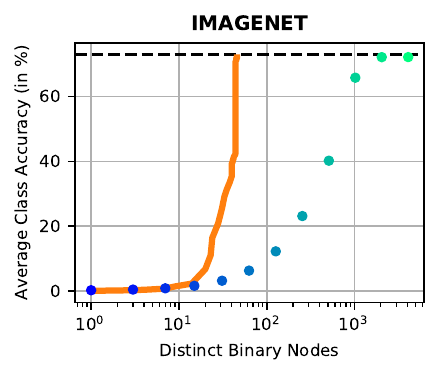}
        \end{center}
    \end{subfigure}
    \begin{subfigure}{0.06\textwidth}
        \includegraphics[width=\textwidth]{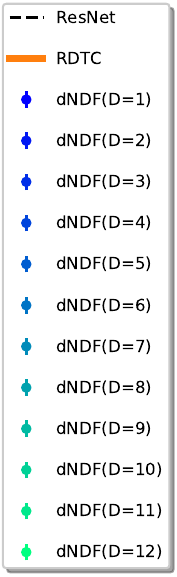}
    \end{subfigure}
    \vspace{-4mm}
    \caption{Accuracy with increasing number of nodes in \texttt{RDTC} and dNDF on AWA2, CUB and ImageNet. 
    As \texttt{RDTC} can reuse learned nodes in the tree and has adaptive tree depth, we train it once and evaluate it at different depths. dNDF needs to be retrained for every depth hyperparameter (D) and the number of nodes scales exponentially with tree depth. }
    \label{fig:ioc_vs_ndf}
    \vspace{-0.5em}
\end{figure*}

Furthermore, we measure the effect of \(\lambda\) to the attribute prediction accuracy of the decision tree as compared to their ground-truth (purple curve). We observe a high attribute prediction accuracy even with a small \(\lambda\), e.g. \(\lambda=0.2\). As we increase \(\lambda\) in the range of 0.2 to 1.0, there is only a slight increase in attribute prediction accuracy, indicating that our \texttt{aRDTC} is robust against the choice of \(\lambda\) across datasets as long as it is chosen to be at least 0.2.

\begin{figure*}[t]
    \centering
    \includegraphics[width=0.9\linewidth]{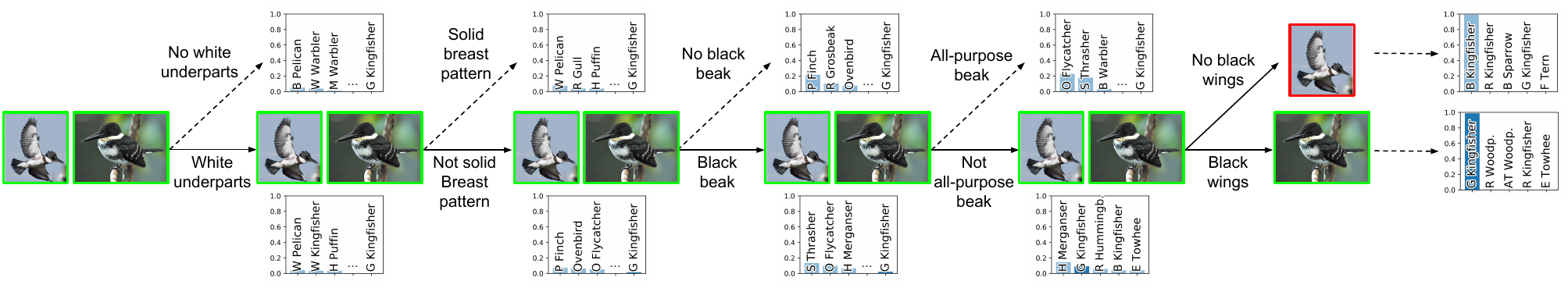}
    \includegraphics[width=0.9\linewidth]{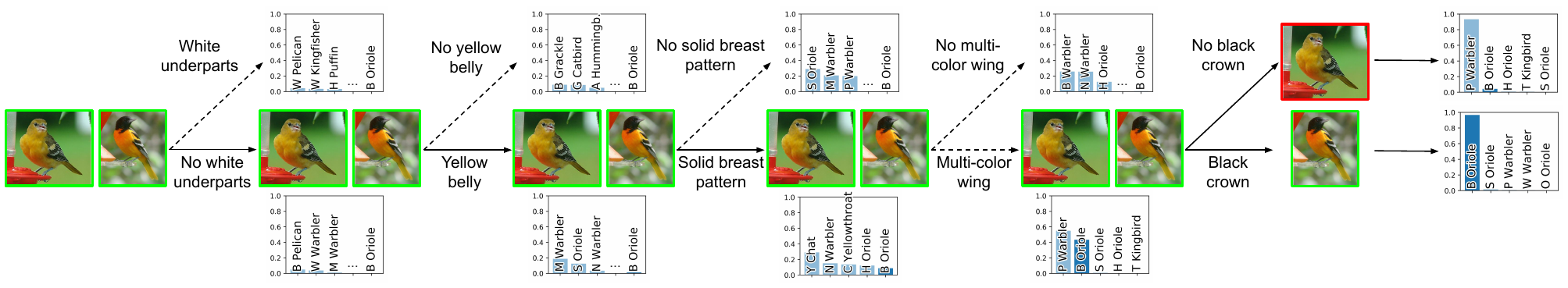}
    \includegraphics[width=0.9\linewidth]{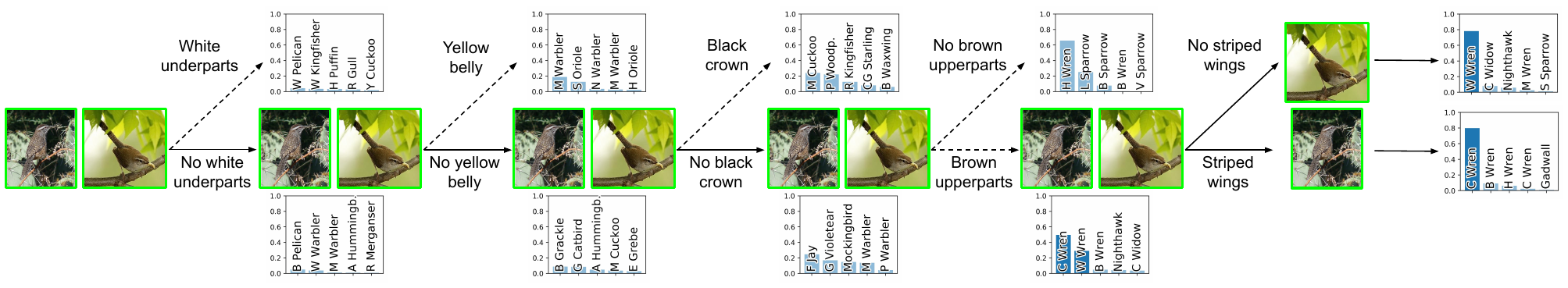}
    \vspace{-2mm}
    \caption{Top: Two ``Green Kingfisher'' images follow the same path except for ``black wings'', i.e. the flying bird gets misclassified as a ``belted kingfisher'' as black wings are not visible. Middle: Baltimore Oriole image (left) gets incorrectly classified as Prothonotary Warbler because of the missing ``black crown'' in the female bird. Such discrepancies, e.g. per-class attributes not reflecting the image content, make CUB difficult.
    Bottom: Cactus Wren (left) and Bewick Wren (right) share many characteristics except from ``striped wings'' which our model uses to split these classes.}
    \label{fig:qual_cub}
\end{figure*}

\myparagraph{Ablating the Memory Mechanism}
The LSTM state \(h\) and explicit memory \(\mathcal{M}\) in RDT contains previously observed decision nodes and the current decision. While the LSTM state allows to encode the attribute order, the explicit memory serves as a more direct representation of all gathered information about the image. We ablate our RDT model with respect to its tree encoding-types. 

Table~\ref{tab:cls_acc_ablation} shows the classification accuracy of the following configurations: \texttt{aRDTC-L}, i.e. with only LSTM and attribute loss, and \texttt{aRDTC-M}, i.e. with only explicit memory, (vs \texttt{RDTC-L} and \texttt{RDTC-M} without the attribute loss). We observe that \texttt{aRDTC-M} consistently performs better than \texttt{aRDTC-L}, e.g. on CUB (76.4\% vs. 69.1\%). Moreover, combining the two in our full model generally improves the performance (up to 1.5\% on CUB).
These results indicate that the explicit memory is important for accuracy.

The median number of distinct attributes in paranthesis shows that \texttt{aRDTC-L} retains fewer decision nodes than \texttt{aRDTC-M}. For instance, \texttt{aRDTC} learns to only use 38 out of all 312 attributes of CUB (\(\approx 12\%\)) and on ImageNet \texttt{RDTC} uses only 46 learned binary attributes as opposed to the 1000 continuous features commonly used in ResNet. This increases sparsity of our model in the attribute space and improves interpretability by using fewer nodes when using the LSTM.
We conclude that combining the two memory types in our RDTC model provides the best of both worlds, a high classification accuracy in few binary decisions such that the explanations of our model are concise and accurate.

\myparagraph{Scalability of the Learned Decision Trees.}
For our \texttt{RDTC}, increasing the tree depth simply translates to increasing the number of binary decisions, i.e., time steps of the model-to-model communication. Hence, \texttt{RDTC} scales linearly with the depth of the tree while the number of weights stays constant. On the other hand, DT and dNDF grow exponentially in their number of parameters with the depth of the tree.
When the same attribute is needed at different locations in the tree, our model learns the meaning of this attribute once and reuses it, while DT and dNDF would have to relearn the split.
Finally, \texttt{RDTC} does not require finetuning a depth parameter.
Hence, we have the flexibility of changing the tree depth at test time without retraining.

We compare the classification accuracy of \texttt{RDTC} and dNDF with an increasing number of distinct tree nodes on three datasets. As shown in Figure~\ref{fig:ioc_vs_ndf}, \texttt{RDTC} (orange line) is trained only once and evaluated at different tree depths at test time while we have to retrain dNDF for each depth parameter. While the number of nodes of dNDF scales exponentially with depth (note the log-scale on the x-axis), our model adaptively learns the number of binary attributes needed to solve these classification tasks. Hence, it stops using more attributes when no further distinct nodes are necessary.
We observe that \texttt{RDTC} uses up to an order of magnitude fewer tree nodes on AWA2, CUB and ImageNet to achieve the same or better performance. At the same time, \texttt{RDTC} only needs to be trained once and can be adaptively reduced in tree depth at test time.

\subsection{Qualitative Results}
\label{sec:visu}

Zooming into the decision process of misclassifications on CUB, we investigate how our model treats counterfactual classes which is useful as explanations are often contrastive~\cite{hendricks2018grounding}. We provide further qualitative examples revealing the decision tree of our aRDTC model on the fine-grained CUB and the decision tree of our RDTC model on ImageNet without the attributes in the supplementary.

In Figure~\ref{fig:qual_cub} (Top), we inspect the point in the tree where the error occurred.
The lower path corresponds to the most probable path taken for birds of class Green Kingfisher. Both images follow the same path for four decisions, the error occurs in the fifth decision.  
For the flying bird, our model decides that it ``does not have black wings'' and incorrectly classifies it as a Belted Kingfisher, a closely related class to Green Kingfisher, but without black wings. In addition, our model depicts its current belief of the correct class at any time during the process, i.e., probability plots at every branch which reveals some critical binary decisions, when the predicted class changes drastically, such as the ``black wings'' decision.
This way, a user inspecting our explainable decision tree can make a more informed decision on the value of the prediction of the model.

In Figure~\ref{fig:qual_cub} (Middle), the Baltimore Oriole image on the left gets incorrectly classified as Prothonotary Warbler because of the missing male-specific ``black crown'' attribute in the female bird. Such discrepancies, e.g. per-class attributes not reflecting the image content, make CUB an extremely challenging dataset. In Figure~\ref{fig:qual_cub} (Bottom), the Cactus Wren image on the left and Bewick Wren image on the right share many characteristics except from ``striped wings''. The decision path is common until then where our model uses this attribute to split these classes.

\section{Conclusion}
In this work, we propose to learn a decision tree recurrently through communication between two-agents. 
Our \texttt{RDTC} framework adaptively changes tree depth at test time, allows to reuse of the learned decision nodes and improves scalability. It also uses human understandable attributes and hard binary splits for easier interpretation. Our experiments show that combining an explicit memory and an LSTM is important to obtain good performances with few inquiries.
Our model maintains the accuracy of non-explainable deep models and outperforming the state-of-the-art deep decision tree learners.
Qualitatively inspecting individual examples demonstrates the reasoning behind the failure and other challenging fine-grained cases, while a user study shows that \texttt{RDTC} selects more visually relevant attributes than a comparable linear semantic bottleneck model.

\myparagraph{Acknowledgements}
This work has been partially funded by the ERC (853489 - DEXIM) and by the DFG (2064/1 – Project number 390727645).

{\small
\bibliographystyle{ieee_fullname}
\bibliography{bib}
}

\twocolumn[
\begin{center}
  \Large\textbf{Learning Decision Trees Recurrently Through Communication\\-\\Supplementary Material}
\end{center}
\vspace{20mm}]
\appendix

\section{Ablating Loss and Maximum Steps T}

When training our \texttt{RDTC} model with a cross-entropy loss for image classification, we found that training is more efficient and yields better results when applying the loss term at every step $t$ in the communication loop up to the maximum step $T$.
\begin{equation}
    \mathcal{L} = \frac{1}{T} \sum_{t=1}^{T} \mathcal{L}_{CE}(y, \hat{y}^{(t)})
    \label{eq:supp_clsloss}
\end{equation} 

\begin{figure}[t]
    \centering
    \includegraphics[width=.8\linewidth]{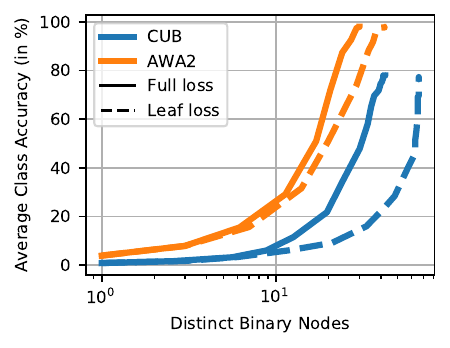}
    \caption{Accuracy and number of distinct nodes in \texttt{RDTC} on AWA2, CUB comparing our full loss at each time step (solid line) with a loss only applied at leaf nodes (dashed line). The full loss uses fewer nodes, i.e., a smaller tree, to achieve the same accuracy.}
    \label{fig:loss_ablation}
    \vspace{-2mm}
\end{figure}

One natural alternative to this approach is to apply the loss only at step $T$, the leaf node, essentially removing the sum Equation~\ref{eq:supp_clsloss}. In Figure~\ref{fig:loss_ablation}, we show the difference of applying the loss at every time step (full loss) or only at the end (leaf loss) on CUB and AWA2. The final performance of the decision tree is the same, however, applying the loss at every time step produces a tree that has a better performance when evaluated at intermediate steps and results in a smaller tree after pruning, i.e., fewer tree nodes are used for the final tree to obtain best performance.

\begin{figure}[t]
    \centering
    \includegraphics[width=.75\linewidth]{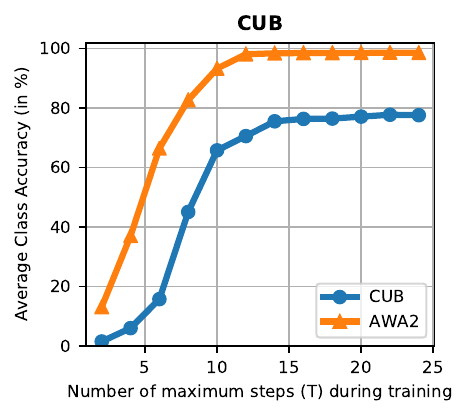}
    \caption{Training \texttt{RDTC} while varying hyperparameter $T$. As $T$ increases, the model achieve a better accuracy up to a value of $T$ where a plateau is reached. When increasing $T$ further, the final tree size of \texttt{RDTC} does not increase due to pruning during tree distillation.}
    \label{fig:depth_ablation}
\end{figure}

Moreover, we found that when hyperparameter $T$ is chosen sufficiently high, we are able to reach this maximum performance while our tree distillation process ensures that the tree size does not increase past the point where the classifier achieves the highest accuracy. Figure~\ref{fig:depth_ablation} shows classification accuracy with increasing tree depth. Accuracy does not decrease past some value for $T$ where the model performs best, and choosing any value bigger results in an equally explainable tree after pruning.

\section{Decision Trees and Explanations of CUB and AWA2}

Illustrating the decision making process helps the user get an explainable overview of the internal decision process of the whole classifier. We point to the tree branch into which a certain class (indicated by an example image from this class) falls along with the attribute associated with that branch.
We inspect the learned structure of the decision tree by illustrating the splits from our \texttt{aRDTC} model on CUB in Figure~\ref{fig:tree_cub}, and on AWA2 in Figure~\ref{fig:tree_awa2}.
Here, the left and right sub-tree indicates that the attribute is present or absent respectively. 
For instance, on CUB, the first decision deals with identifying bird with white underparts, separating these from birds with any other color.
These categories get further refined with each binary split via a hierarchical clustering that reveals the decision tree structure of our \texttt{aRDTC} framework.
These serve as additional examples of introspection, showing that our model allows to make a more informed decision about the trustworthiness of the network's prediction.

In Figure~\ref{fig:qual_cub}, we illustrate a qualitative example of the classification of two images of Scarlet Tanagers made by our model trained on CUB. Both images follow the same path for the first decisions, before diverging when it comes to the decision whether the bird has black wings. The top bird actually does not have black wings and, thus, is classified as a Summer Tanager, a bird species with the same appearance as Scarlet Tanager except for having red instead of black wings.

Equivalently in Figure~\ref{fig:qual_awa2}, we illustrate a qualitative example of the classification of two images of tigers made by our model trained on AWA2. Again, both images follow the same decision until, for the white tiger, our model wrongly predicts ``no stripes'' and incorrectly classifies it as a lion. 
Together with the full decision trees, these explanations allow for detailed introspections into the global decision process our \texttt{aRDTC} model.

\section{Explanations without Attributes}

\begin{figure}[t]
    \centering
    \includegraphics[width=\linewidth, trim=0 0 0 0, clip]{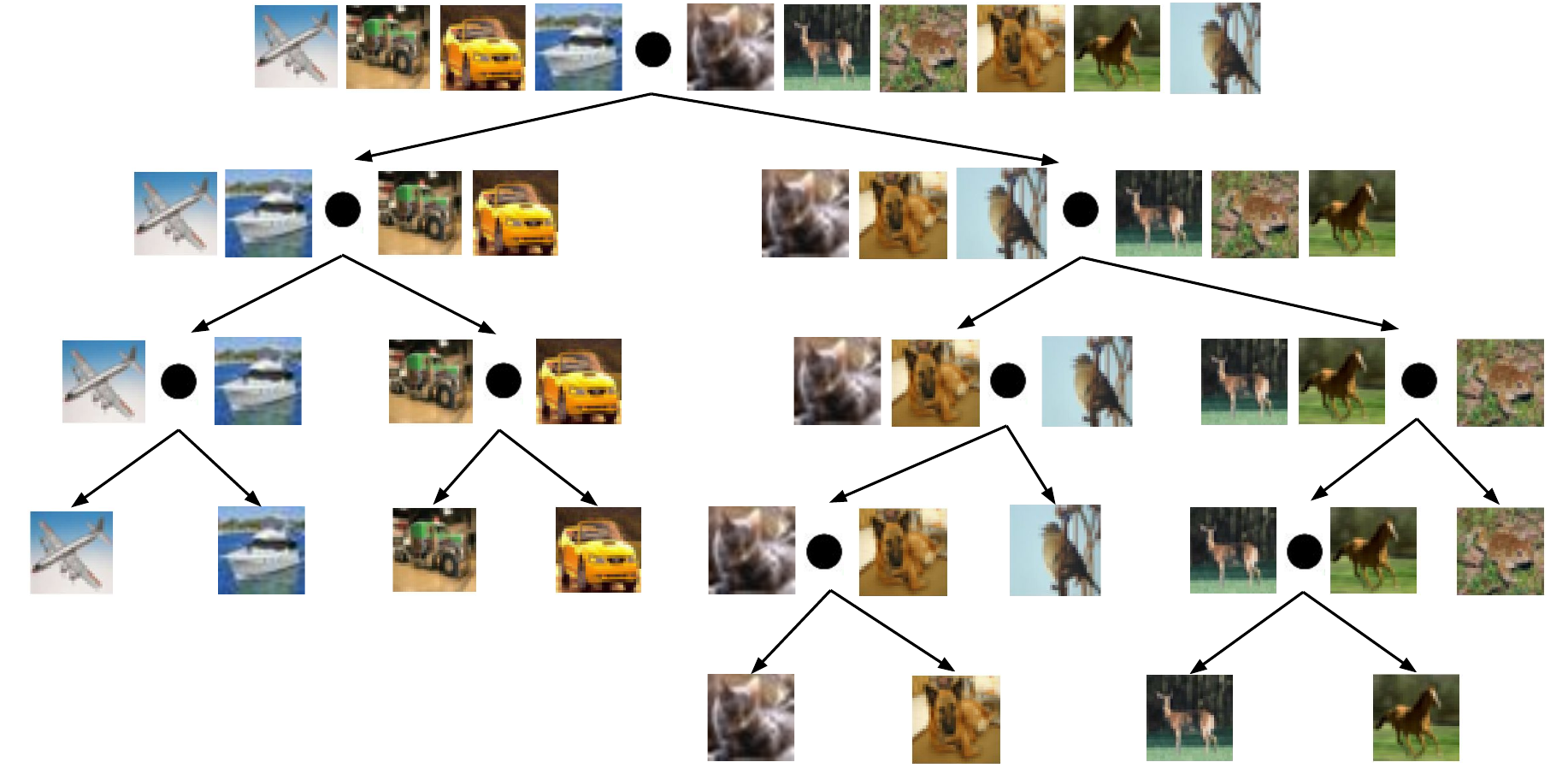}
    \caption{Our \texttt{RDTC} learns the decision tree on CIFAR10 without attribute data. While decision nodes do not have ground truth attributes, we can still interpret the decision, e.g., the first node separates animals from vehicles.}
    \label{fig:tree_cifar10}
\end{figure}

When working with datasets that do not provide annotated attributes, we can train our \texttt{RDTC} with \(\lambda=0\), which still exposes the decision tree structure. This allows introspection into the intermediate class splits of the model revealing a hierarchy that can reveal semantics.
When applies on CIFAR-10, our \texttt{RDTC} model not only retains ResNet performance ($93.1\%$ vs. $93.3\%$), it also semantically clusters the data even though there is no attribute guidance. Figure~\ref{fig:tree_cifar10} shows the resulting decision tree of \texttt{RDTC} on CIFAR-10. In the first binary split, we observe that \texttt{RDTC} separates the animal classes from the vehicles. Subsequently, vehicles are clustered into motor vehicles (car, truck) and the rest (airplane, ship). For animals, our model also finds reasonable clusters such as grouping cat and dog, as well as grouping horse and deer.
ImageNet is a more challenging dataset, where we observe similar behaviour. In Figure~\ref{fig:tree_imagenet}, we show the decision tree of the first decisions on ImageNet with a randomly selected subset of classes, each represented by one representative image. Our model separates animals from inanimate objects in the first tree split following the data semantics. In the later decisions of the tree, there are clusters of dogs/cats, birds, monkeys on one side of the tree and clusters of furniture and electrical appliances on the other. These example show that, even when no additional attribute information is given, tree splits often follow semantics that are exposed by the decision tree learned by our \texttt{RDTC}.

\section{RDTC Training Algorithm}
For a concise representation of the \texttt{RDTC} training algotithm, we present a summary in Algorithm~\ref{alg:train} including both components, RDT and AbL, iterative loss calculation and gradient updates using the terminology of the main paper.

\begin{figure}[t]
\begin{minipage}{0.49\textwidth}
\begin{algorithm}[H]
\begin{algorithmic}[1]
    \REQUIRE{Image \(x\), label \(y\)\\
             \hspace{2.2em}Max \# of decisions \(T\), Attribute data \(\alpha\) }
    \ENSURE{Predicted label \(\hat{y}\)\\
            \hspace{2.6em}Binary decision sequence \(d^{(1)}, \dots, d^{(T)}\)}
    \STATE \(z =\) CNN(\(x\))
    \STATE \(\boldsymbol{\hat{a}} = \text{TempSoftmax}(f_{\text{AttrMLP}}(z))\)
    \STATE init \(\mathcal{M}^{0}, h_0\)
    \STATE \(\mathcal{L} = 0\)
    \FOR{\(t=1\) to \(T\)}
        \STATE \(c^{(t)} = \text{GumbelSoftmax}(f_{\text{QuestMLP}}(h^{(t-1)}))\)
	    \STATE \(d^{(t)} = \hat{a}_{c^{(t)}}\)
	    \STATE \(\mathcal{M}^{(t)} = \mathcal{M}^{(t-1)} \oplus (c^{(t)}, d^{(t)})\)
	    \STATE \(h^{(t)} = \text{LSTM}(h^{(t-1)}, \mathcal{M}^{(t)}, c^{(t)}, d^{(t)})\)
	    \STATE \(\hat{y}^{(t)} = f_{\text{ClassMLP}}(\mathcal{M}^{(t)})\)
	    \STATE \(\mathcal{L}^{(t)} = \frac{1}{T} \Big[(1-\lambda) \mathcal{L}_{CE}(y, \hat{y}^{(t)}) + \lambda \mathcal{L}_{CE}(\alpha_{y,c^{(t)}}, \hat{a}_{c^{(t)}})\Big]\)
	    \STATE \(\mathcal{L} = \mathcal{L} + \mathcal{L}^{(t)}\)
    \ENDFOR
    \STATE gradient update with \(\mathcal{L}\)
    \STATE \textbf{return} \(\hat{y}^{(T)}; d^{(1)}, \dots, d^{(T)}\)
\end{algorithmic}
\caption{RDTC training}
\label{alg:train}
\end{algorithm}
\end{minipage}
\end{figure}

\onecolumn
\begin{landscape}
\begin{figure*}
    \begin{minipage}{\linewidth}
        \centering
        \includegraphics[width=\linewidth, trim=0 0 0 0, clip]{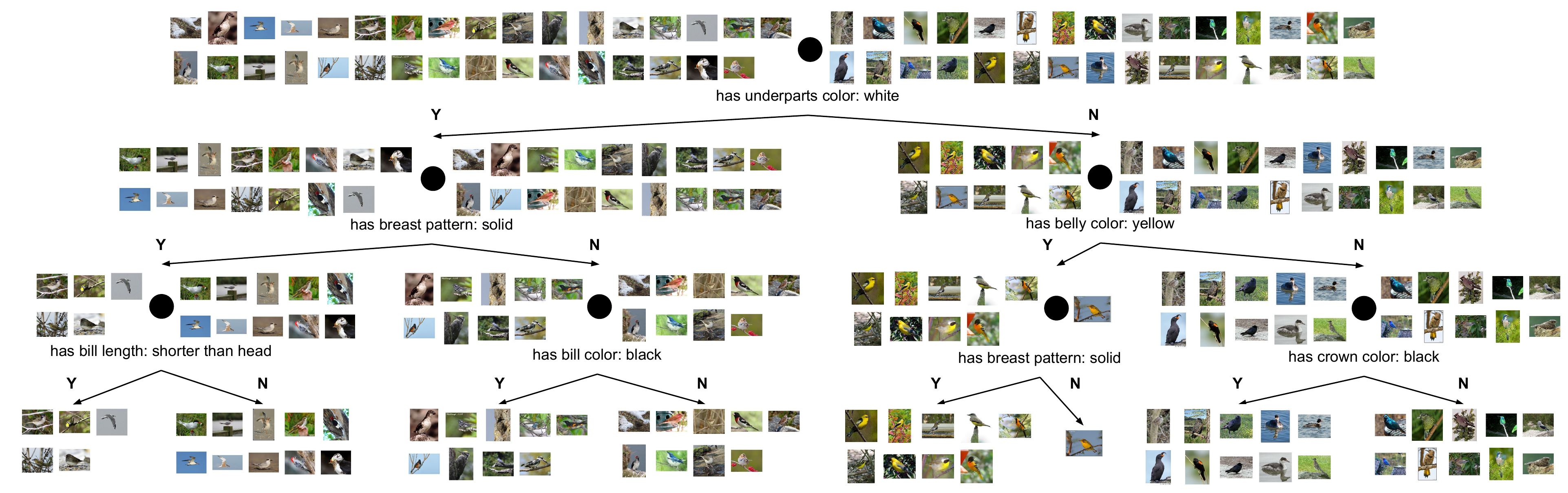}
        \caption{Our \texttt{aRDTC} learns explainable decisions via the decision tree showing the path for each class and it also gives each decision a human-understandable meaning. Here we show the first three decisions for a subset of the 200 classes of birds in CUB where a randomly selected image from a class represents each class. }
        \label{fig:tree_cub}
    \end{minipage}
    \begin{minipage}{\linewidth}
        \centering
        \includegraphics[width=\linewidth]{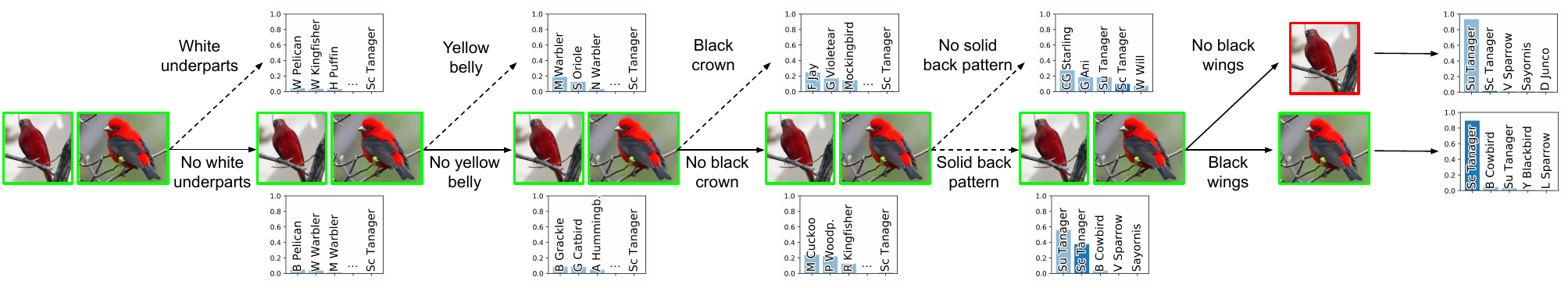}
        \caption{Our \texttt{aRDTC} points to the reasoning behind a wrong decision. Here we illustrate two images from the ``Green Kingfisher'' class. The lower path lead to a correct classification. Both images follow the same path except for the decision of ``black wings''. The flying bird gets classified as a ``Belted Kingfisher'' incorrectly because the black wings are not visible.}
        \label{fig:qual_cub_supp}
    \end{minipage}
\end{figure*}

\begin{figure*}
    \begin{minipage}{\linewidth}
        \centering
        \includegraphics[width=\linewidth, trim=0 0 0 0, clip]{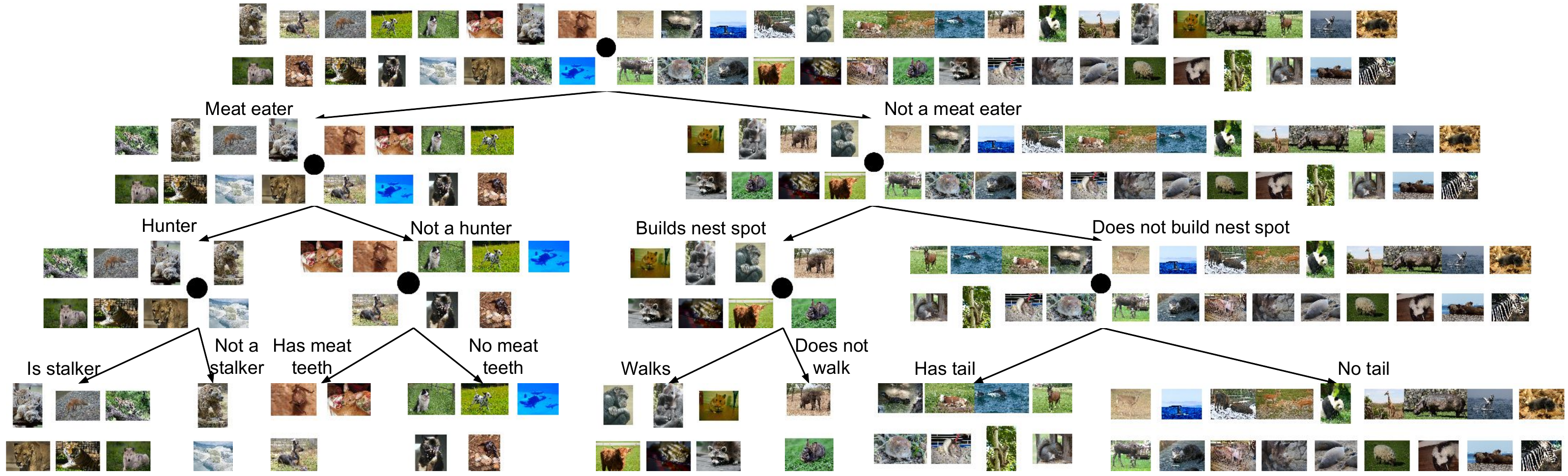}
        \caption{Learned explainable decisions on AWA2 by our aRDTC model. We show the decision tree of the most likely path for each class, \emph{i.e.}, introspection, and give each decision a human-understandable meaning, \emph{i.e.}, rationalization. The tree exposes the thought process of our model, \emph{e.g.}, it decides to separate meat-eating animals from all other animals in the first step.}
        \label{fig:tree_awa2}
    \end{minipage}
    \begin{minipage}{\linewidth}
        \centering
        \includegraphics[width=\linewidth]{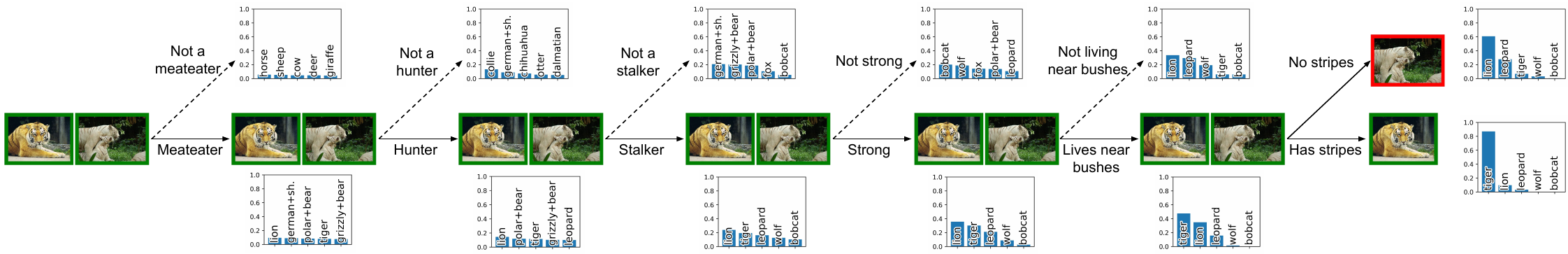}
        \caption{Decision process for two tiger images in AWA2 along with the current label prediction at each step. The lower (upper) path is taken when the attribute is present (absent) for a given class. Both images follow the same path except for the last decision, ``has stripes''. Since these are absent in the white tiger, it gets classified incorrectly as a lion.}
        \label{fig:qual_awa2}
    \end{minipage}
\end{figure*}

\begin{figure*}
    \centering
    \includegraphics[width=\linewidth, trim=0 0 0 0, clip]{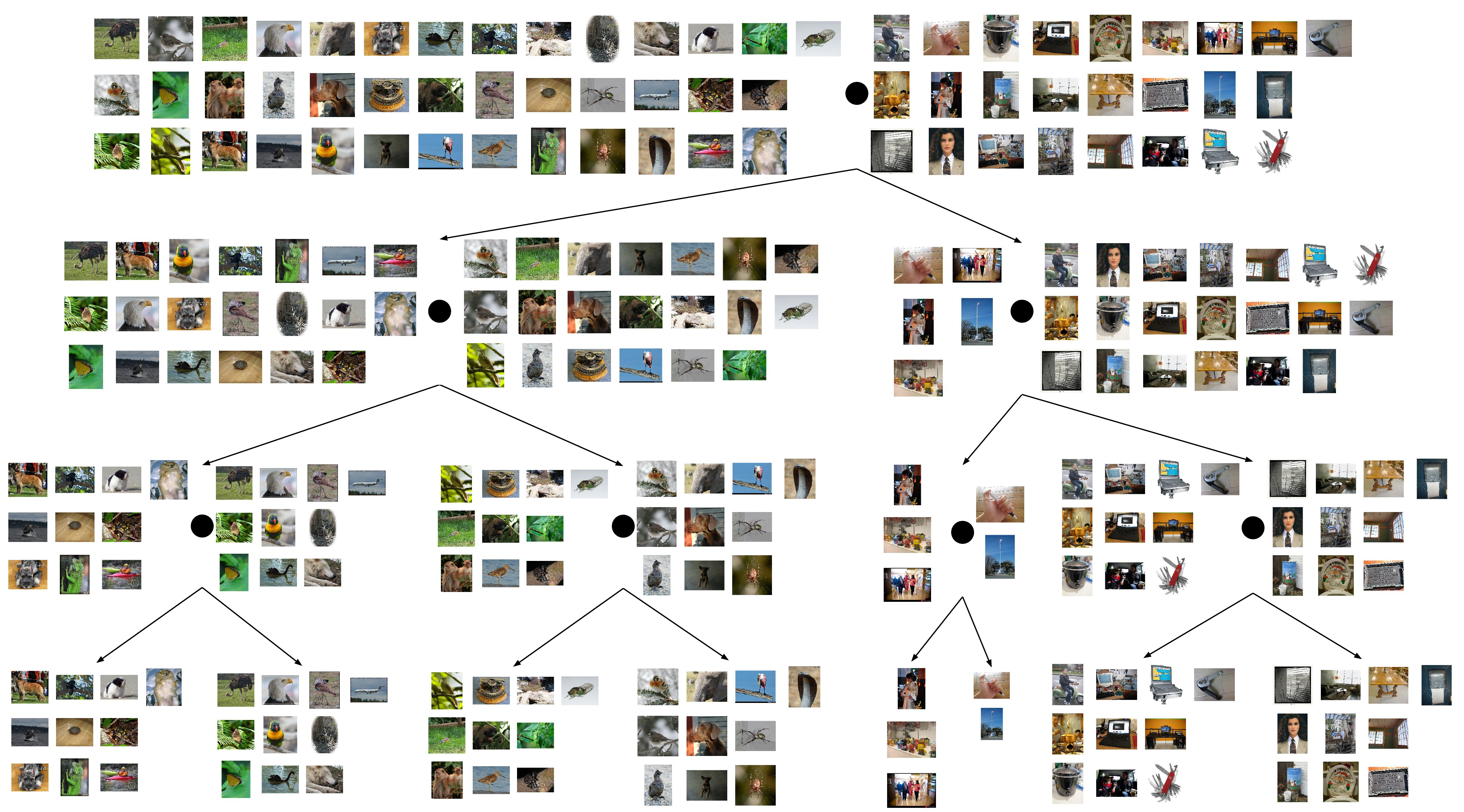}
    \caption{Learned binary decisions on ImageNet by our RDTC model. A subset of randomly chosen classes are shown by one representative image of each class. Our tree reveals a clustering as decision splits narrow down towards a specific subset of classes.}
    \label{fig:tree_imagenet}
\end{figure*}
\end{landscape}

\end{document}